\documentclass[conference]{IEEEtran}
\IEEEoverridecommandlockouts
\usepackage{cite}
\usepackage{amsmath,amssymb,amsfonts}
\usepackage{algorithmic}
\usepackage{graphicx}
\usepackage{textcomp}
\usepackage{xcolor}
\usepackage{multirow}
\usepackage{subfigure}
\usepackage{tabularx}
\usepackage{booktabs}
\def\BibTeX{{\rm B\kern-.05em{\sc i\kern-.025em b}\kern-.08em
    T\kern-.1667em\lower.7ex\hbox{E}\kern-.125emX}}
\begin{document}

\title{A Distillation-based Future-aware Graph Neural Network for Stock Trend Prediction\\
\thanks{* Corresponding author.}
}

\author{\IEEEauthorblockN{Zhipeng Liu}
\IEEEauthorblockA{\textit{Software college} \\
\textit{Northeastern University}\\
Shenyang, China\\
2310543@stu.neu.edu.cn
}
\and
\IEEEauthorblockN{Peibo Duan$^*$}
\IEEEauthorblockA{\textit{Software college} \\
\textit{Northeastern University}\\
Shenyang, China\\
sakuragiduan@gmail.com}
\and
\IEEEauthorblockN{Mingyang Geng}
\IEEEauthorblockA{\textit{Unit 63798} \\
Xichang, China \\
gengmingyang13@nudt.edu.cn}
\and
\IEEEauthorblockN{Bin Zhang}
\IEEEauthorblockA{\textit{Software college} \\
\textit{Northeastern University}\\
Shenyang, China \\
zhangbin@mail.neu.edu.cn}
}

\maketitle

\begin{abstract}
Stock trend prediction involves forecasting the future price movements by analyzing historical data and various market indicators. With the advancement of machine learning, graph neural networks (GNNs) have been extensively employed in stock prediction due to their powerful capability to capture spatiotemporal dependencies of stocks. However, despite the efforts of various GNN stock predictors to enhance predictive performance, the improvements remain limited, as they focus solely on analyzing historical spatiotemporal dependencies, overlooking the correlation between historical and future patterns. In this study, we propose a novel distillation-based future-aware GNN framework (DishFT-GNN) for stock trend prediction. Specifically, DishFT-GNN trains a teacher model and a student model, iteratively. The teacher model learns to capture the correlation between distribution shifts of historical and future data, which is then utilized as intermediate supervision to guide the student model to learn future-aware spatiotemporal embeddings for accurate prediction. Through extensive experiments on two real-world datasets, we verify the state-of-the-art performance of DishFT-GNN.
\end{abstract}


\section{Introduction and Related work}
\label{sec:intro}
The stock market is a financial investment platform where numerous corporations and investors engage in trading\cite{hu2015application,liu2024dynamic}. As of 2022, the global market capitalization increased from \$54.6 trillion to \$94.69 trillion over the past decade\footnote{  https://data.worldbank.org/indicator/CM.MKT.LCAP.CD/}. With the advancement of artificial intelligence, deep learning (DL) techniques provide more opportunities for investors to increase their wealth through stock investment\cite{deep_stock_predict1,htun2023survey}. For an auxiliary investment tool, it is non-trivial to make profitable investment decisions and trading strategies within the volatile market\cite{ding2024trend}. 

\begin{figure}
  \centering
  \includegraphics[scale=.23]{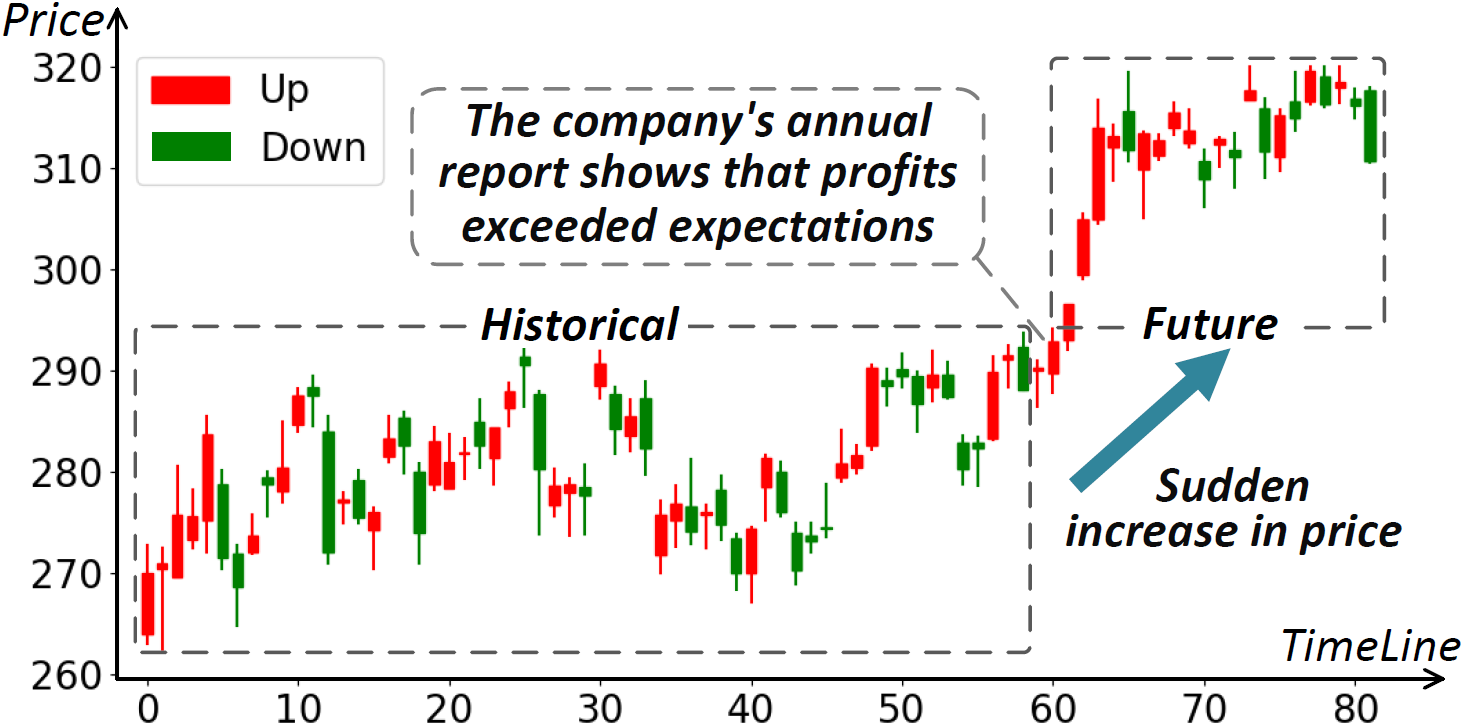}
  \caption{An example shows that events, such as a corporation’s annual report, cause sudden structural changes between historical and future stock data.}
  \label{intro}
\end{figure}
Typically, DL-based stock trading methods are based on the fundamental hypothesis that a stock's \textbf{future dynamics} can be revealed from its \textbf{historical patterns}\cite{trans_sp2,trans_sp3,hist1,hist2,wang2022adaptive}. Thus, it is essential to perform a meticulous analysis of latent temporal dynamics within historical stock price indicators, as exemplified by methods based on recurrent neural networks (RNN) \cite{2015_lstmsp_chen,2018_lstmsp_Fischer}, as well as its variants LSTM or GRU \cite{2017_DARNN_Qin,2017_attention_Vaswani,2019_alstm_feng}. Drawing upon the temporal dynamic feature extraction capabilities of RNN-based methodologies, graph neural networks (GNNs) construct stock graphs that facilitate the incorporation of explicitly relational (spatial) dependencies between stocks, such as industry and shareholding information \cite{TGC,ye_2021_multi,hist,qian2024mdgnn}. In this framework, the nodes denote individual stocks with attributes derived from the aforementioned RNN-based models, while the edges represent the inter-stock relationships. Recent studies have found that relying solely on explicit relationships can lead to biased and incomplete aggregation of relational features. As a result, there is growing interest in investigating implicit relationships using DL-based graph representation methods to better understand stock interactions\cite{ADGAT,2023_mgar_song,vgnn,yan2024framework,wang2024towards}.

Going beyond the above mentioned studies, we further observe that stock data exhibits highly non-stationary characteristics, making it more challenging to forecast compared to other time series data. As shown in Fig.\ref{intro}, an annual report from a corporation showing better-than-expected profits can cause distribution shift between historical and future data, which results in sub-optimal performance in contemporary stock predictors that rely exclusively on historical price indicators, with their impact becoming apparent only in hindsight. Thus, we are motivated to believe that not only are the features in historical stock data essential for analysis, but also \textbf{the correlation between distribution shifts of historical and future data}. For further proof, please refer to the supplementary material provided due to space constraints\footnote{https://github.com/ZhupengLiu/DisFT-GNN/blob/main/Supplementary\%20\\Material.pdf}.

To this end, we propose a novel general \textbf{Dis}tillation-based \textbf{F}u\textbf{T}ure-aware \textbf{GNN} framework (\textbf{DisFT-GNN}), to capture the correlation between historical and future patterns for stock trend prediction. In theory, DisFT-GNN can be integrated with any GNN-based stock predictor, enhancing their predictive performance. The contributions in this study are summarized as follows:\textbf{ i)} Unlike existing knowledge distillation-based time series forecasting methods\cite{distill1,distill2}, which utilize the teacher model to produce historical pattern-related soft labels or features for guiding the training of a lightweight student model, DisFT-GNN introduces a novel teacher model that has undergone converged training to generate high-level, future-aware representations, which serve as intermediate supervision, guiding the student model in learning correlations between historical and future distributions. \textbf{ii) }As the future prices can be influenced by various factors and exhibit different patterns. We introduce a novel attention-based multi-channel feature fusion method in the teacher model, which models the diverse distribution shifts between historical and future data. \textbf{iii)} Through extensive experiments on two real-world datasets from the American stock market, our proposed DisFT-GNN consistently boosts current state-of-the-art models, achieving an average improvement of up to 5.41\%.
    




\section{Problem Formulation}
We formulate stock trend prediction as a binary node classification task. A dynamic stock graph at trading day $t$ can be defined as $\mathcal{G}^t=\{\mathcal{V},\mathbf{X}^{[t-L+1,t]},\mathbf{A}^t\}$, where $\mathcal{V}=\{v_1,v_2,...,v_N\}$ is the set of $N$ individual stocks, $\mathbf{X}^{[t-L+1,t]} \in \mathbb{R}^{N \times L \times M}$ represents $M$ price indicators over the historical $L$ trading days. $\mathbf{A}^t\in \mathbb{R}^{N\times N}$ is the normalized relation matrix representing the relationship between stocks. Mathematically, the problem is formulated as:
\begin{equation}
    \hat{\mathbf{y}}^{[t+1,t+T]}=f(g(\mathbf{X}^{[t-L+1,t]},\mathbf{A}^t;\Theta_g);\Theta_f).
    \label{eq: problem formulation}
\end{equation}

\noindent In \eqref{eq: problem formulation}, $g(\cdot)$ is a GNN-based model with parameters $\Theta_g$, aiming to capture spatiotemporal features between stocks; $f(\cdot)$ is the prediction layer with parameters $\Theta_f$; $\hat{\mathbf{y}}^{[t+1,t+T]} = \{\hat{\mathbf{y}}_1^{[t+1,t+T]},\hat{\mathbf{y}}_2^{[t+1,t+T]},..., \hat{\mathbf{y}}_N^{[t+1,t+T]} \}$ is the set of output binary variables. For $\forall \tau \in [t+1,t+T]$, $\hat{y}_n^{\tau} = 1$ predicts that the $n$-th stock will increase, and 0 otherwise. For $\forall v_n\in \mathcal{V}$, the ground-truth label is defined as follows:
\begin{equation}
    y_n^{\tau}=
    \begin{cases}
        1& \text{if }\frac{p_n^\tau-p_n^{t}}{p_n^{t}}>\delta\\
        0& \text{else }
    \end{cases},
\end{equation}
\noindent where $p_n^ \tau\in \mathbf{x}_n^\tau$ is the stock close price at trading day $\tau$, $\delta$ is the hyperparameter. Notably, when $T=1$ and $\delta=0$, the problem is the next-trading day stock trend prediction \cite{you2024multi}.

\begin{figure}
  \centering
  \includegraphics[scale=.27]{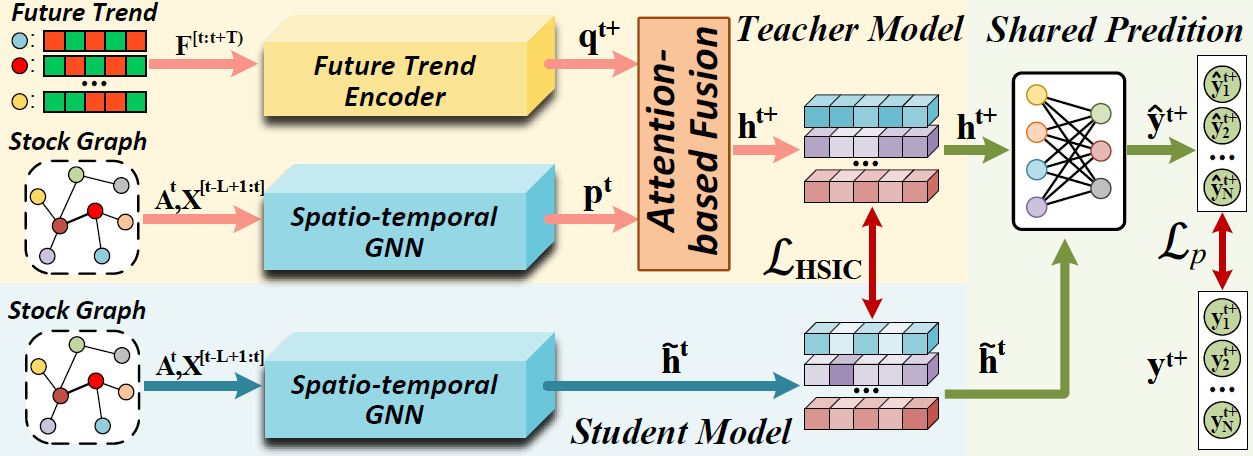}
  \caption{The framework of DishFT-GNN}
  \label{framework}
\end{figure}

\section{Methodology}
\subsection{Overall Architecture}
The overview of DisFT-GNN is illustrated in Fig.\ref{framework}. DisFT-GNN contains two training processes: training a teacher model and a student model, respectively. First, the teacher model encodes the historical stock graph and future trend information, generating historical spatiotemporal and high-level future embeddings. Subsequently, a novel attention-based multi-channel feature fusion method is proposed to integrate these embeddings to generate future-aware spatiotemporal representations, which depict the diverse historical-future distribution shifts. The representations are then fed into the prediction module to obtain predicted results, which are compared with the ground truth to optimize the teacher model. Subsequently, the future-aware spatiotemporal representations from the teacher model serve as intermediate supervision to guide the student model in learning historical-future distribution correlations.

Before introducing the details of the proposed method, we first introduce several general components for stock prediction.

\noindent{\textbf{Spatiotemporal GNN module.}} Existing stock prediction methods follow a fundamental hypothesis that a stock's future prices can be inferred from its historical patterns (temporal dependence)\cite{2017_DARNN_Qin} and the behavior of related stocks (spatial dependence)\cite{ADGAT}. For $\forall v_n\in\mathcal{V}$, the spatiotemporal embeddings can be calculated as $\mathbf{h}_n^t=ST(\mathbf{X}_n^{[t-L+1,t]}, \mathbf{A}^t)$, specifically,
\begin{equation}
    \begin{aligned}
\mathbf{s}_n^t&=\text{Temporal}(\mathbf{X}_n^{[t-L+1,t]}),\\
\mathbf{h}_n^t&=\text{Spatial}(\mathbf{s}_k^t|v_k\in \mathcal{N}_n),
    \end{aligned} 
\end{equation}

\noindent where $\mathcal{N}_n$ denotes the set of stocks related to $v_n$. It is noted that $ST(\cdot)$ can be any spatiotemporal GNN model.

\noindent{\textbf{Future Trend Encoder.}} It is a feed-forward network (FNN) to encode the stock future trends, generating a novel future trend embedding, ${\mathbf{q}}_n^{t+}$. Since future trend information can intuitively reflect prediction results, it is unnecessary to design intricate modules to analyze future price indicators.

\begin{equation}
    {\mathbf{q}}_n^{t+}=\text{ReLU}(\text{FFN}(\mathbf{f}_n^{[t:t+T)})),
\end{equation}
\noindent where $\mathbf{f}_n^{[t:t+T)}\in\{0, 1\}$ represents the future trend of $v_n$ over the following $T$ trading days.




\subsection{Future-aware Teacher Model Training}
\subsubsection{Future-aware Spatiotemporal Encoding}
First, the teacher model utilizes a future trend encoder that encodes stock future trends into novel high-level embeddings, ${\mathbf{q}}_n^{t+}=\text{ReLU}(\text{FFN}(\mathbf{f}_n^{[t:t+T)})) $, ${\mathbf{q}}_n^{t+}\in \mathbb{R}^{D_f}$, and a spatiotemporal GNN module generating historical spatiotemporal embeddings, $\mathbf{p}_n^{t}=ST_{(T)}(\mathbf{X}_n^{[t-L+1,t]}, \mathbf{A}^t)$, $\mathbf{p}_n^{t}\in \mathbb{R}^{D_p}$. Subsequently, to model the diverse historical-future distribution shifts, we propose a novel attention-based multi-channel fusion method, integrating ${\mathbf{q}}_n^{t+}$ and $\mathbf{p}_n^t$ into high-level future-aware spatiotemporal representations, $\mathbf{h}_n^{t+}\in \mathbb{R}^D$.

Specifically, we perform multiple vector-matrix-vector (VMV) multiplications \cite{ADGAT} to generate a multi-channel result, with each channel representing a potential correlation between historical and future distributions, $V=\mathbf{p}_n^t \mathcal{F}^{[1:{D}]} \mathbf{q}_n^{t+}$, where $\mathcal{F}^{[1:{D}]} \in \mathbb{R}^{D\times D_p \times D_f}$ is the trainable parameter, and $D$ is the number of channels. Next, we utilize the attention mechanism to assess the importance of each channel by assigning an attention score, i.e., $Q={\mathbf{q}}_n^{t+} W_Q$, $K=\mathbf{p}_n^t W_K$. Finally, similar to Transformer\cite{vaswani2017attention}, we conduct a scaled dot-product attention operation for feature fusion, which is formulated as,
\begin{equation}
\mathbf{h}_n^{t+}=\text{Attention}(Q,K,V)=\text{Softmax}(\tau\frac{QK^T}{\sqrt{K_d}})V,
\end{equation}

\noindent where Softmax($\cdot$) is the softmax function, ${K_d}$ is the scaling factor and $\tau$ is  the attention temperature coeffcient.

\subsubsection{Prediction and Optimization}
After obtaining $\mathbf{h}_n^{t+}$, to forecast the stock trend, $\mathbf{h}_n^{t+}$ is fed into the prediction module, which is a feed-forward network with a softmax function.
\begin{equation}
    \hat{\mathbf{y}}_n^{[t+1,t+T]}=\text{Prediction}(\mathbf{h}_n^{t+})=\text{Softmax}(\text{FNN}(\mathbf{h}_n^{t+}))
\end{equation}

\noindent where $\hat{\mathbf{y}}_n^{[t+1,t+T]}\in \{0,1\}$ is the predicted result and the softmax function generates a probability distribution over classes. 

Finally, parameters are learned by minimizing the cross entropy loss (CEL), $\mathcal{L}_p=\text{CEL}({\mathbf{y}}_n^{[t+1,t+T]},\hat{\mathbf{y}}_n^{[t+1,t+T]})$.

\subsection{Distillation-based Student Model Training}
To enable the student model with the capability to deduct the future patterns based on historical spatiotemporal embeddings, we utilize the teacher model, which has undergone converged training, to supervise the training of the student model, which includes two phases.

First, the student model encodes the historical stock graph, generating historical spatiotemporal representations, $\tilde{\mathbf{h}}_n^{t}=ST_{(S)}(\mathbf{X}_n^{[t-L+1,t]}, \mathbf{A}^t)$. Subsequently, $\tilde{\mathbf{h}}_n^t$ is fed into the shared prediction module with the same parameters as the teacher model for prediction,

\begin{equation}
    \hat{\mathbf{y}}_n^{[t+1,t+T]}=\text{Prediction}(\tilde{\mathbf{h}}_n^t).
\end{equation}

Second, to infer future patterns based on $\tilde{\mathbf{h}}^{t}_n$, the student model is trained by minimizing the distillation loss between $\tilde{\mathbf{h}}^{t}_n$ and $\mathbf{h}_n^{t+}$: $\mathcal{L}_{d}=\text{MSE}(\tilde{\mathbf{h}}^{t}_n,\mathbf{h}_n^{t+})$, where MSE($\cdot$) is the mean square error function and $\mathbf{h}_n^{t+}$ is the future-aware spatiotemporal representation distilled from the teacher model. Note that $\mathbf{h}_n^{t+}$ and $\tilde{\mathbf{h}}^{t}_n$ have the same dimensions. However, since the representations are highly non-linear, it is necessary to focus more on the nonlinear dependence between $\tilde{\mathbf{h}}^{t}_n$ and $\mathbf{h}_n^{t+}$. Thus, the Hilbert-Schmidt Independence Criterion (HSIC)\cite{fan2023generalizing,nag2021graphvicreghsic} is utilized as the distillation loss,

\begin{equation}
    \mathcal{L}_{d}=\text{HSIC}(\tilde{\mathbf{h}}^{t}_n,\mathbf{h}_n^{t+})={(D-1)}^{-2}tr(\mathbf{K}\mathbf{H}\mathbf{L}\mathbf{H}),
\end{equation}
\noindent where $\mathbf{K},\mathbf{L}\in\mathbb{R}^{D\times D}$ are kernel matrices, and $\mathbf{K}_{i,j}=k(\tilde{\mathbf{h}}^{t}_i,\tilde{\mathbf{h}}^{t}_j)$, $\mathbf{L}_{i,j}=l(\tilde{\mathbf{h}}^{t}_i,\tilde{\mathbf{h}}^{t}_j)$. $\mathbf{H}=\mathbf{I}-D^{-1}\mathbf{11}^T$, where $\mathbf{I}$ is an identity matrix and $\mathbf{1}$ is an all-one column vector. 

The final objective of the student model is to generate future patterns based on the historical stock data while forecasting the stock trends as accurately as possible. Specifically,
\begin{equation}
    \mathcal{L}=\mathcal{L}_p+\lambda \mathcal{L}_d,
\end{equation}
\noindent where $\mathcal{L}_p=\text{CEL}({\mathbf{y}}_n^{[t+1,t+T]},\hat{\mathbf{y}}_n^{[t+1,t+T]})$, and $\lambda$ is the hyperparameter striking a balance between two terms. 


\section{experiments}
\subsection{Dataset and Experimental Setting}
\subsubsection{Datasets} To verify the effectiveness of the proposed DishFT-GNN, we conduct extensive experiments using two datasets from American stock indices, i.e., S\&P 100 and NASDAQ 100, spanning from January 1, 2019, to September 30, 2023. All the datasets are divided into three parts: 85\% for training, 7.5\% for validation and 7.5\% for testing. To ensure that both datasets contain continuous trading records, we eliminate stocks with missing data, such as those affected by suspensions. Thus, 96 and 94 stocks are selected from S\&P 100 and NASDAQ 100, respectively. Additionally, industry data is utilized to represent explicit stock relationships.

\subsubsection{Experimental Setting} 

To assess the performance of DishFT-GNN, we conduct the experiments with seven GNN stock prediction methods for comparison, i.e., GCN\cite{kipf2016semi}, GAT\cite{velickovic2017graph}, TGC\cite{TGC}, ADGAT\cite{ADGAT}, MGAR\cite{2023_mgar_song}, VGNN\cite{vgnn} and MDGNN\cite{qian2024mdgnn}.  DishFT-GNN is integrated with each stock predictor to boost predictive performance. Following previous studies\cite{Macro-Sector-Micro,pen}, we use accuracy (ACC) and Matthews correlation coefficient (MCC) as two metrics.

Parameters of all models are trained using Adam optimizer\cite{kingma2014adam} on a single NVIDIA RTX 4070Ti GPU. In our experiments, $T$ and $\delta$ are set to 20 and 4\%, respectively. $\tau$ is set to 0.5 The learning rate is set to 5e-4 and the batch size is set to 64. We independently repeated each experiment five times and reported the mean and standard deviation.


\begin{table}[t]
\scriptsize
    \centering
    \caption{Performance Comparison with Baselines.}
    \begin{tabular}{ccccc}
    \toprule[1pt]
\multirow{2}{*}{}
   \multirow{2}{*}{Methods} &\multicolumn{2}{c}{S\&P 100}&\multicolumn{2}{c}{NASDAQ 100}\\
   \cmidrule(lr){2-3}\cmidrule(lr){4-5}
    & ACC(\%) & MCC & ACC(\%) & MCC \\
    \midrule[.5pt]
        GCN&52.12$\pm$1.57&0.050$\pm$0.008& 51.97$\pm$0.84&0.026$\pm$0.005\\
   +Dish-FT&\textbf{55.47}$\pm$\textbf{1.18}&\textbf{0.159}$\pm$\textbf{0.035}&\textbf{53.87}$\pm$\textbf{1.26}&\textbf{0.093}$\pm$\textbf{0.013}\\
    \midrule[.5pt]
        GAT&51.81$\pm$1.25&0.047$\pm$0.011&52.14$\pm$1.34&0.048$\pm$0.008\\
   +Dish-FT&\textbf{55.45}$\pm$\textbf{1.23}&\textbf{0.092}$\pm$\textbf{0.013}&\textbf{54.38}$\pm$\textbf{1.27}&\textbf{0.097}$\pm$\textbf{0.017}\\
    \midrule[.5pt]
        TGC&52.84$\pm$0.85&0.059$\pm$0.007&53.61$\pm$1.30&0.082$\pm$0.016\\
   +Dish-FT&\textbf{56.69}$\pm$\textbf{0.78}&\textbf{0.137}$\pm$\textbf{0.012}&\textbf{57.13}$\pm$\textbf{1.21}&\textbf{0.140}$\pm$\textbf{0.043}\\
    \midrule[.5pt]
      ADGAT&53.46$\pm$1.34&0.077$\pm$0.023& 53.39$\pm$1.60&0.063$\pm$0.012\\
   +Dish-FT&\textbf{57.71}$\pm$\textbf{1.62}&\textbf{0.167}$\pm$\textbf{0.024}& \textbf{57.82}$\pm$\textbf{2.86}&\textbf{0.145}$\pm$\textbf{0.008}\\
    \midrule[.5pt]
       MGAR&52.93$\pm$0.62&0.051$\pm$0.008& 52.78$\pm$1.34&0.079$\pm$0.021\\
   +Dish-FT&\textbf{56.26}$\pm$\textbf{1.59}&\textbf{0.145}$\pm$\textbf{0.027}& \textbf{58.19}$\pm$\textbf{1.87}&\textbf{0.162}$\pm$\textbf{0.037}\\
    \midrule[.5pt]
       VGNN&53.73$\pm$1.48&0.075$\pm$0.014& 53.91$\pm$1.48&0.079$\pm$0.025\\
   +Dish-FT&\textbf{58.31}$\pm$\textbf{1.34}&\textbf{0.160}$\pm$\textbf{0.015}& \textbf{58.26}$\pm$\textbf{1.27}&\textbf{0.156}$\pm$\textbf{0.044}\\
    \midrule[.5pt]
      MDGNN&52.52$\pm$0.36&0.051$\pm$0.012& 52.84$\pm$1.19&0.058$\pm$0.018\\
   +Dish-FT&\textbf{55.68}$\pm$\textbf{0.55}&\textbf{0.093}$\pm$\textbf{0.018}& \textbf{56.53}$\pm$\textbf{1.05}&\textbf{0.127}$\pm$\textbf{0.013}\\
    
\bottomrule[1pt]
    \end{tabular}
    \label{result}
\end{table}

\subsection{Results and Analysis}

Table \ref{result} illustrates the comparison results, where each row is divided into two sections: the upper section displays the predictive performance of the original baseline methods, while the lower section presents the performance achieved after applying the proposed DishFT-GNN. Additionally, we apply a t-test at a significance level of $\alpha=0.01$ to validate the reliability and reproducibility of DishFT-GNN\cite{boneau1960effects}. We observe that DishFT-GNN significantly ($p<0.01$) enhances predictive performance compared to all baseline models. Specifically, it achieves an accuracy improvement over 2\% in most cases and up to 5.41\% when integrating MGAR, and an MCC improvement of 82.3\% to 218\% across two datasets.

\begin{figure}[t]
  \scriptsize
  \subfigure[S\&P 100.]{\includegraphics[width=0.22\textwidth]{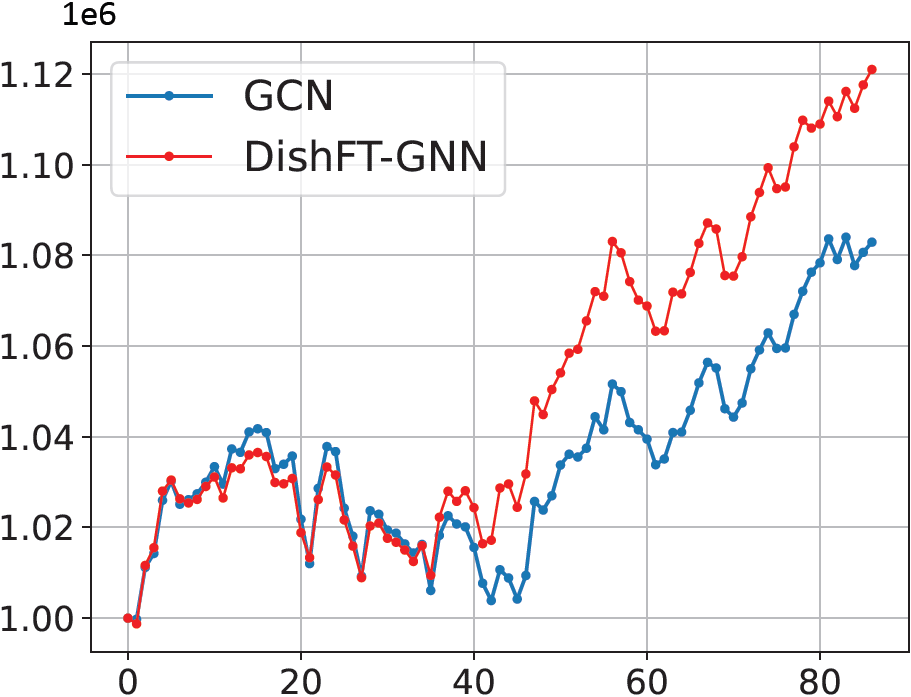}}
\hspace{0.01cm}
  \subfigure[NASDAQ 100.]{\includegraphics[width=0.22\textwidth]{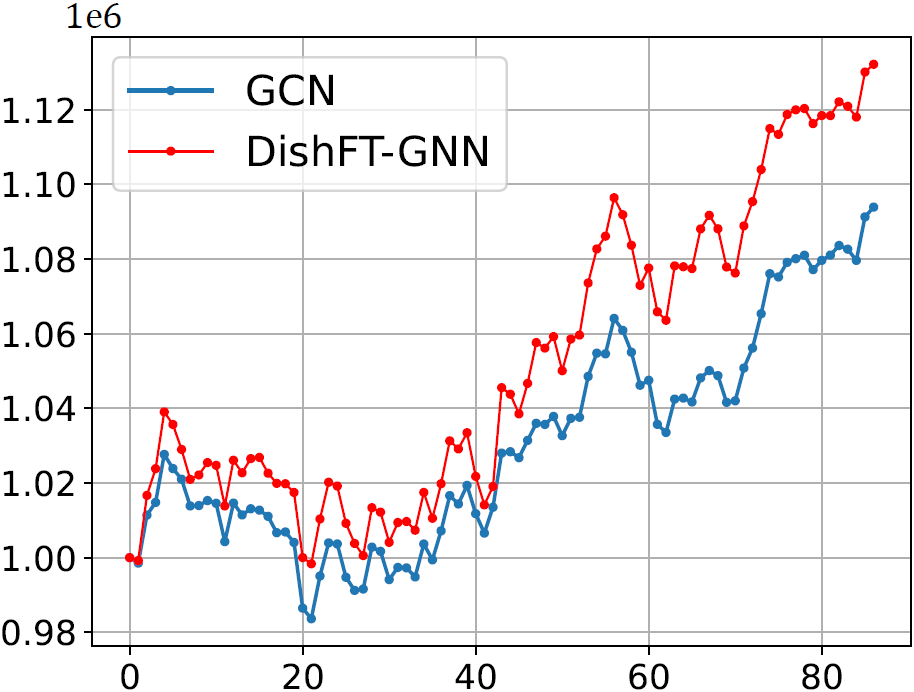}}
  \caption{Back-testing illustration.}
  \label{Backtesting}
\end{figure}

Additionally, to evaluate whether the proposed DishFT-GNN improves model profitability, we use the classic GCN as the backbone and compare the investment profits with and without DishFT-GNN for back-testing, which is illustrated in Fig.\ref{Backtesting}. It is obvious that the profit gap between two methods gradually widens over time, which further verify the effectiveness of DishFT-GNN.

Previous studies rely solely on historical stock knowledge, which is insufficient for forecasting future outcomes. Despite capturing comprehensive historical spatiotemporal dependencies, these models can only achieve sub-optimal improvements. Since historical data only reflect past stock states, analyzing them alone may fail to capture future stock dynamics in a volatile market environment. Therefore, based on the motivation and theoretical proof in Section \ref{sec:intro}, the proposed DishFT-GNN utilizes the teacher model to explore the correlations between historical-future distribution shifts and supervises the student model's training, which enables the student model to analyze future stock patterns based on historical spatiotemporal knowledge, effectively alleviate the problem of the model's inability to establish historical-future associations. Notably, during the back-testing stage, only the student model is deployed, ensuring that no future information is leaked.

\begin{table}[t]
\scriptsize
    \centering
    \caption{Ablation study.}
    \begin{tabularx}{.42\textwidth}{cc|cc|cc}
    \toprule[1pt]
   \multicolumn{2}{c}{\multirow{2}{*}{Methods}} &\multicolumn{2}{c}{S\&P 100} &\multicolumn{2}{c}{NASDAQ 100} \\
   \cmidrule(lr){3-4}\cmidrule(lr){5-6}
   && ACC(\%)& MCC&ACC(\%)& MCC \\
   \midrule[.5pt]
   \multirow{3}{*}{\rotatebox{90}{GCN}}
   &DishFT-GNN w/o H&54.38&0.082&52.98&0.071\\
   &DishFT-GNN w/o F&54.78&0.075&52.18&0.092\\
   &DishFT-GNN &\textbf{55.47}&\textbf{0.159}&\textbf{53.87}&\textbf{0.093}\\
    \midrule[.5pt]
   \multirow{3}{*}{\rotatebox{90}{GAT}}
   &DishFT-GNN w/o H&53.81&0.058&52.42&0.063\\
   &DishFT-GNN w/o F&54.54&0.072&53.45&0.052\\
   &DishFT-GNN &\textbf{55.45}&\textbf{0.092}&\textbf{54.38}&\textbf{0.097}\\
       \midrule[.5pt]
   \multirow{3}{*}{\rotatebox{90}{VGNN}}
   &DishFT-GNN w/o H&55.34&0.108&56.21&0.115\\
   &DishFT-GNN w/o F&56.17&0.123&57.28&0.137\\
   &DishFT-GNN &\textbf{58.31}&\textbf{0.160}&\textbf{58.26}&\textbf{0.156}\\
    \midrule[.5pt]
   \multirow{3}{*}{\rotatebox{90}{MDGNN}}
   &DishFT-GNN w/o H&54.72&0.088&54.51&0.074\\
   &DishFT-GNN w/o F&55.68&0.076&55.04&0.091\\
   &DishFT-GNN &\textbf{55.68}&\textbf{0.093}&\textbf{56.53}&\textbf{0.127}\\
\bottomrule[1pt]
    \end{tabularx}
    \label{ablation}
\end{table}

\subsection{Ablation Study}
To investigate the contributions of each component, we conduct the ablation study on two classic baselines, GCN and GAT, and two state-of-the-art baselines, VGNN and MDGNN.

\begin{itemize}
    \item \textbf{DishFT-GNN w/o H:} It replaces HSIC with MSE.
    \item \textbf{DishFT-GNN w/o F:} It replaces attention-based multi-channel feature fusion module with simple concatenation.
    \item \textbf{DishFT-GNN:} Vanilla DishFT-GNN framework.
\end{itemize}

\noindent The ablation results are illustrated in Table \ref{ablation}. We can observe that all components positively impact the performance of DishFT-GNN in terms of ACC and MCC. First, HSIC aids the student model in learning a broader range of nonlinear features. Moreover, the attention-based multi-channel fusion method effectively models diverse historical-future distribution shifts, leading to improved predictive performance.

\section{conclusion}

We propose a novel distillation-based framework named DishFT-GNN for stock trend prediction. To address the issue that contemporary stock prediction methods fail to extract rich future patterns from historical stock knowledge, we introduced a teacher model trained to generate informative historical-future correlations, supervising the student model's learning. Specifically, the teacher model first extracted historical spatiotemporal and future embeddings. We then proposed a novel attention-based multi-channel fusion method to mine the association between historical and future distributions by integrating both embeddings. Finally, the fused representation was then utilized as intermediate supervision to guide the student GNN to learn future-aware spatiotemporal representations for accurate prediction. Extensive evaluations on real-world data showcase the effectiveness of our proposed method.

\section{Acknowledgements}
This study was supported by the Department of Science and Technology of Liaoning province (02110076523001), and Northeastern University, Shenyang, China (02110022124005).

\bibliographystyle{IEEEbib}
\bibliography{refs}

\begin{thebibliography}{10}

\bibitem{hu2015application}
Yong Hu, Kang Liu, Xiangzhou Zhang, Lijun Su, EWT Ngai, and Mei Liu,
\newblock ``Application of evolutionary computation for rule discovery in stock
  algorithmic trading: A literature review,''
\newblock {\em Applied Soft Computing}, vol. 36, pp. 534--551, 2015.

\bibitem{liu2024dynamic}
Zhipeng Liu, Peibo Duan, Xiaosha Xue, Changsheng Zhang, Wenwei Yue, and Bin
  Zhang,
\newblock ``A dynamic hypergraph attention network: Capturing market-wide
  spatiotemporal dependencies for stock selection,''
\newblock {\em Applied Soft Computing}, p. 112524, 2024.

\bibitem{deep_stock_predict1}
Xiao Ding, Yue Zhang, Ting Liu, and Junwen Duan,
\newblock ``Deep learning for event-driven stock prediction,''
\newblock in {\em Twenty-fourth international joint conference on artificial
  intelligence}, 2015.

\bibitem{htun2023survey}
Htet~Htet Htun, Michael Biehl, and Nicolai Petkov,
\newblock ``Survey of feature selection and extraction techniques for stock
  market prediction,''
\newblock {\em Financial Innovation}, vol. 9, no. 1, pp. 26, 2023.

\bibitem{ding2024trend}
Wei Ding, Zhennan Chen, Hanpeng Jiang, Yuanguo Lin, and Fan Lin,
\newblock ``Trend-heuristic reinforcement learning framework for news-oriented
  stock portfolio management,''
\newblock in {\em ICASSP 2024-2024 IEEE International Conference on Acoustics,
  Speech and Signal Processing (ICASSP)}. IEEE, 2024, pp. 5120--5124.

\bibitem{trans_sp2}
Qianggang Ding, Sifan Wu, Hao Sun, Jiadong Guo, and Jian Guo,
\newblock ``Hierarchical multi-scale gaussian transformer for stock movement
  prediction.,''
\newblock in {\em IJCAI}, 2020, pp. 4640--4646.

\bibitem{trans_sp3}
Qiuyue Zhang, Chao Qin, Yunfeng Zhang, Fangxun Bao, Caiming Zhang, and Peide
  Liu,
\newblock ``Transformer-based attention network for stock movement
  prediction,''
\newblock {\em Expert Systems with Applications}, vol. 202, pp. 117239, 2022.

\bibitem{hist1}
Yumo Xu and Shay~B Cohen,
\newblock ``Stock movement prediction from tweets and historical prices,''
\newblock in {\em Proceedings of the 56th Annual Meeting of the Association for
  Computational Linguistics (Volume 1: Long Papers)}, 2018, pp. 1970--1979.

\bibitem{hist2}
Huihui Ni, Shuting Wang, and Peng Cheng,
\newblock ``A hybrid approach for stock trend prediction based on tweets
  embedding and historical prices,''
\newblock {\em World Wide Web}, vol. 24, no. 3, pp. 849--868, 2021.

\bibitem{wang2022adaptive}
Heyuan Wang, Tengjiao Wang, Shun Li, Jiayi Zheng, Shijie Guan, and Wei Chen,
\newblock ``Adaptive long-short pattern transformer for stock investment
  selection.,''
\newblock in {\em IJCAI}, 2022, pp. 3970--3977.

\bibitem{2015_lstmsp_chen}
Kai Chen, Yi~Zhou, and Fangyan Dai,
\newblock ``A lstm-based method for stock returns prediction: A case study of
  china stock market,''
\newblock in {\em 2015 IEEE international conference on big data (big data)}.
  IEEE, 2015, pp. 2823--2824.

\bibitem{2018_lstmsp_Fischer}
Thomas Fischer and Christopher Krauss,
\newblock ``Deep learning with long short-term memory networks for financial
  market predictions,''
\newblock {\em European journal of operational research}, vol. 270, no. 2, pp.
  654--669, 2018.

\bibitem{2017_DARNN_Qin}
Yao Qin, Dongjin Song, Haifeng Chen, Wei Cheng, Guofei Jiang, and Garrison
  Cottrell,
\newblock ``A dual-stage attention-based recurrent neural network for time
  series prediction,''
\newblock {\em arXiv preprint arXiv:1704.02971}, 2017.

\bibitem{2017_attention_Vaswani}
Ashish Vaswani, Noam Shazeer, Niki Parmar, Jakob Uszkoreit, Llion Jones,
  Aidan~N Gomez, {\L}ukasz Kaiser, and Illia Polosukhin,
\newblock ``Attention is all you need,''
\newblock {\em Advances in neural information processing systems}, vol. 30,
  2017.

\bibitem{2019_alstm_feng}
Fuli Feng, Huimin Chen, Xiangnan He, Jie Ding, Maosong Sun, and Tat-Seng Chua,
\newblock ``Enhancing stock movement prediction with adversarial training.,''
\newblock in {\em IJCAI}, 2019, vol.~19, pp. 5843--5849.

\bibitem{TGC}
Fuli Feng, Xiangnan He, Xiang Wang, Cheng Luo, Yiqun Liu, and Tat-Seng Chua,
\newblock ``Temporal relational ranking for stock prediction,''
\newblock {\em ACM Transactions on Information Systems (TOIS)}, vol. 37, no. 2,
  pp. 1--30, 2019.

\bibitem{ye_2021_multi}
Jiexia Ye, Juanjuan Zhao, Kejiang Ye, and Chengzhong Xu,
\newblock ``Multi-graph convolutional network for relationship-driven stock
  movement prediction,''
\newblock in {\em 2020 25th International Conference on Pattern Recognition
  (ICPR)}. IEEE, 2021, pp. 6702--6709.

\bibitem{hist}
Wentao Xu, Weiqing Liu, Lewen Wang, Yingce Xia, Jiang Bian, Jian Yin, and
  Tie-Yan Liu,
\newblock ``Hist: A graph-based framework for stock trend forecasting via
  mining concept-oriented shared information,''
\newblock {\em arXiv preprint arXiv:2110.13716}, 2021.

\bibitem{qian2024mdgnn}
Hao Qian, Hongting Zhou, Qian Zhao, Hao Chen, Hongxiang Yao, Jingwei Wang, Ziqi
  Liu, Fei Yu, Zhiqiang Zhang, and Jun Zhou,
\newblock ``Mdgnn: Multi-relational dynamic graph neural network for
  comprehensive and dynamic stock investment prediction,''
\newblock {\em arXiv preprint arXiv:2402.06633}, 2024.

\bibitem{ADGAT}
Rui Cheng and Qing Li,
\newblock ``Modeling the momentum spillover effect for stock prediction via
  attribute-driven graph attention networks,''
\newblock in {\em Proceedings of the AAAI conference on artificial
  intelligence}, 2021, vol.~35, pp. 55--62.

\bibitem{2023_mgar_song}
Guowei Song, Tianlong Zhao, Suwei Wang, Hua Wang, and Xuemei Li,
\newblock ``Stock ranking prediction using a graph aggregation network based on
  stock price and stock relationship information,''
\newblock {\em Information Sciences}, vol. 643, pp. 119236, 2023.

\bibitem{vgnn}
Rong Xing, Rui Cheng, Jiwen Huang, Qing Li, and Jingmei Zhao,
\newblock ``Learning to understand the vague graph for stock prediction with
  momentum spillovers,''
\newblock {\em IEEE Transactions on Knowledge and Data Engineering}, 2023.

\bibitem{yan2024framework}
Yuxiao Yan, Changsheng Zhang, Xiaohang Li, and Bin Zhang,
\newblock ``A framework for stock selection via concept-oriented attention
  representation in hypergraph neural network,''
\newblock {\em Knowledge-Based Systems}, vol. 284, pp. 111326, 2024.

\bibitem{wang2024towards}
Binwu Wang, Pengkun Wang, Yudong Zhang, Xu~Wang, Zhengyang Zhou, Lei Bai, and
  Yang Wang,
\newblock ``Towards dynamic spatial-temporal graph learning: A decoupled
  perspective,''
\newblock in {\em Proceedings of the AAAI Conference on Artificial
  Intelligence}, 2024, vol.~38, pp. 9089--9097.

\bibitem{distill1}
Lei Li, Zhiyuan Zhang, Ruihan Bao, Keiko Harimoto, and Xu~Sun,
\newblock ``Distributional correlation--aware knowledge distillation for stock
  trading volume prediction,''
\newblock in {\em Joint European Conference on Machine Learning and Knowledge
  Discovery in Databases}. Springer, 2022, pp. 105--120.

\bibitem{distill2}
Qing Xu, Zhenghua Chen, Mohamed Ragab, Chao Wang, Min Wu, and Xiaoli Li,
\newblock ``Contrastive adversarial knowledge distillation for deep model
  compression in time-series regression tasks,''
\newblock {\em Neurocomputing}, vol. 485, pp. 242--251, 2022.

\bibitem{you2024multi}
Zinuo You, Pengju Zhang, Jin Zheng, and John Cartlidge,
\newblock ``Multi-relational graph diffusion neural network with parallel
  retention for stock trends classification,''
\newblock in {\em ICASSP 2024-2024 IEEE International Conference on Acoustics,
  Speech and Signal Processing (ICASSP)}. IEEE, 2024, pp. 6545--6549.

\bibitem{vaswani2017attention}
A~Vaswani,
\newblock ``Attention is all you need,''
\newblock {\em Advances in Neural Information Processing Systems}, 2017.

\bibitem{fan2023generalizing}
Shaohua Fan, Xiao Wang, Chuan Shi, Peng Cui, and Bai Wang,
\newblock ``Generalizing graph neural networks on out-of-distribution graphs,''
\newblock {\em IEEE Transactions on Pattern Analysis and Machine Intelligence},
  2023.

\bibitem{nag2021graphvicreghsic}
Sayan Nag,
\newblock ``Graphvicreghsic: Towards improved self-supervised representation
  learning for graphs with a hyrbid loss function,''
\newblock {\em arXiv preprint arXiv:2105.12247}, 2021.

\bibitem{kipf2016semi}
Thomas~N Kipf and Max Welling,
\newblock ``Semi-supervised classification with graph convolutional networks,''
\newblock {\em arXiv preprint arXiv:1609.02907}, 2016.

\bibitem{velickovic2017graph}
Petar Velickovic, Guillem Cucurull, Arantxa Casanova, Adriana Romero, Pietro
  Lio, Yoshua Bengio, et~al.,
\newblock ``Graph attention networks,''
\newblock {\em stat}, vol. 1050, no. 20, pp. 10--48550, 2017.

\bibitem{Macro-Sector-Micro}
Guifeng Wang, Longbing Cao, Hongke Zhao, Qi~Liu, and Enhong Chen,
\newblock ``Coupling macro-sector-micro financial indicators for learning stock
  representations with less uncertainty,''
\newblock in {\em Proceedings of the AAAI Conference on Artificial
  Intelligence}, 2021, vol.~35, pp. 4418--4426.

\bibitem{pen}
Shuqi Li, Weiheng Liao, Yuhan Chen, and Rui Yan,
\newblock ``Pen: prediction-explanation network to forecast stock price
  movement with better explainability,''
\newblock in {\em Proceedings of the AAAI Conference on Artificial
  Intelligence}, 2023, vol.~37, pp. 5187--5194.

\bibitem{kingma2014adam}
Diederik~P Kingma,
\newblock ``Adam: A method for stochastic optimization,''
\newblock {\em arXiv preprint arXiv:1412.6980}, 2014.

\bibitem{boneau1960effects}
C~Alan Boneau,
\newblock ``The effects of violations of assumptions underlying the t test.,''
\newblock {\em Psychological bulletin}, vol. 57, no. 1, pp. 49, 1960.

\end{thebibliography}

\end{document}